\documentclass[conference]{IEEEtran}
\IEEEoverridecommandlockouts
\usepackage{cite}
\usepackage{amsmath,amssymb,amsfonts}
\usepackage{algorithmic}
\usepackage{graphicx}
\usepackage{textcomp}
\usepackage[linesnumbered,ruled,vlined]{algorithm2e}
\usepackage{xcolor}
\usepackage{pifont}

\usepackage{xspace}
\usepackage{caption}
\usepackage{subcaption}
\usepackage{amsmath}
\usepackage{cite}
\usepackage{bm}
\usepackage{amsmath}
\usepackage{array}
\usepackage{setspace}
\usepackage[nameinlink]{cleveref}

\usepackage{mathtools}
\usepackage{amsthm}
\usepackage{multirow}
\usepackage[linesnumbered,ruled,vlined]{algorithm2e}
\usepackage{tabularx}
\SetKwRepeat{Do}{do}{while}
\usepackage{color}
\usepackage{epstopdf}
\usepackage{endnotes}
\usepackage{framed}

\usepackage{cool} 
\usepackage{enumerate} 

\usepackage{pifont}

\Crefname{figure}{Fig.}{Figs.}

\usepackage[subnum]{cases}

\newcommand{\SumNoLim}[2]{\ensuremath{\sum\nolimits_{#1}^{#2}}}

\usepackage{mathrsfs}
\usepackage[justification=centering]{caption}
\allowdisplaybreaks[4]

\newcolumntype{P}[1]{>{\centering\arraybackslash}p{#1}}
\newcommand{\xmarklight}{\ding{53}}%

\DeclareUnicodeCharacter{2212}{-}

\usepackage{color}

\usepackage{titlesec}
\titlespacing{\section}{0pt}{\parskip}{-\parskip}
\titlespacing{\subsection}{0pt}{\parskip}{-\parskip}

\def\BibTeX{{\rm B\kern-.05em{\sc i\kern-.025em b}\kern-.08em
    T\kern-.1667em\lower.7ex\hbox{E}\kern-.125emX}}
\begin{document}
\title{\LARGE Threshold-Based Data Exclusion Approach for Energy-Efficient Federated Edge Learning}
\author{\IEEEauthorblockN{Abdullatif Albaseer, Mohamed Abdallah, Ala Al-Fuqaha, and~Aiman Erbad}
\IEEEauthorblockA{Division of Information and Computing Technology, College of Science and Engineering,
\\Hamad Bin Khalifa University, Doha, Qatar \\
\{amalbaseer, moabdallah, aalfuqaha, AErbad\}@hbku.edu.qa}
}
\maketitle

\begin{abstract}
Federated edge learning (FEEL) is a promising distributed learning technique for next-generation wireless networks. FEEL preserves the user's privacy, reduces the communication costs, and exploits the unprecedented capabilities of edge devices to train a shared global model by leveraging a massive amount of data generated at the network edge. However, FEEL might significantly shorten energy-constrained participating devices' lifetime due to the power consumed during the model training round. This paper proposes a novel approach that endeavors to minimize computation and communication energy consumption during FEEL rounds to address this issue. First, we introduce a modified local training algorithm that intelligently selects only the samples that enhance the model's quality based on a predetermined threshold probability. Then, the problem is formulated as joint energy minimization and resource allocation optimization problem to obtain the optimal local computation time and the optimal transmission time that minimize the total energy consumption considering the worker's energy budget, available bandwidth, channel states, beamforming, and local CPU speed. After that, we introduce a tractable solution to the formulated problem that ensures the robustness of FEEL. Our simulation results show that our solution substantially outperforms the baseline FEEL algorithm as it reduces the local consumed energy by up to 79\%.
 
\end{abstract}


\IEEEpeerreviewmaketitle

\section{INTRODUCTION}
 The rapid advancement in fifth-generation (5G) cellular networks and the Internet of Things (IoT) have brought a sharp rise in the volume of data generated by end devices (e.g., smartphones, IoT devices, and smart sensors) and cellular base stations (BSs) at the wireless network edge. 
 According to recent research reports, the increase is projected to be in the order of billions of gigabytes per day by 2022~\cite{forecast2019cisco}. 
 This data can bring valuable artificial intelligence (AI) services to end-users by leveraging deep learning~\cite{lecun2015deep} and mobile edge computing techniques~\cite{zhu2020toward}, which have been evolving and converging rapidly in recent years under the umbrella of mobile edge learning. It is envisioned that mobile edge learning will be a core technology for sixth-generation (6G) cellular networks enabling new applications, such as virtual reality and augmented reality; therefore, actualizing the vision of network intelligence\cite{letaief2019roadmap,mo2020energy}. However, transferring large amounts of data to a central unit has become very difficult because of the networks' constraints, scalability issues, limited bandwidth, and most importantly, users' privacy. 

Recently, federated edge learning (\textbf{FEEL}) has shown great potential on the network edge to process data collaboratively among end-users while maintaining privacy since data remains on client devices and only local models are shared~\cite{li2020federated}. 
Also, FEEL significantly reduces the network traffic load, as users only need to share the models' parameters, which are fixed in size and structure across users. 

In FEEL, the convergence of the learning algorithm may not be guaranteed. 
The algorithm's convergence rate determines the number of training rounds required to reach the desired training accuracy. 
The model update tasks in each training round depend on the computation and communications latencies and the energy budget existing at edge workers. 
Therefore, it is clear that training the local models involves computation and communication costs and energy constraints closely interlinked. 
Hence, a scalable FEEL system that guarantees application performance must consider the learning algorithms, the communications system, and the edge devices' energy consumption aspects. 
These aspects impose new challenges that must be considered in designing an efficient FEEL~\cite{zhu2020toward}. 

This paper addresses these issues by proposing a novel scheme for allocating computation and communication resources to support energy-efficient FEEL systems. We study FEEL using a realistic wireless network where the workers are connected to a radio-frequency (RF) base station (BS). Each worker trains its local model using its local data and sends the model parameters to the BS, which aggregates all local parameters to form a global model that is broadcast to all available workers for further updates. Since the BS is resource-constrained, only a subset of available workers is selected every round to perform the updates. Furthermore, the workers are energy-constrained with finite batteries. Thus, the resources must be allocated to workers that meet all the required constraints. The key contributions of this work can be summarized as follows:
\begin{itemize}
    \item We utilize a system's approach that jointly considers the learning algorithms as well as the available computation and communication resources to design an energy-efficient FEEL training algorithm. The proposed algorithm leverages the global model to filter the local data samples and then intelligently selects only the local training samples that improve the model's performance.
    
    \item  We formulate a joint energy minimization and resource allocation problem for FEEL.  This problem is then solved as a sub-problem using a local heuristic that filters the data samples and using the Golden-Section search method to optimize the communication and computation energy.
    
    \item We carry out extensive simulations using realistic federated datasets to assess the proposed solutions' efficacy. Our experiments demonstrate that the proposed technique can reduce the local energy consumption by up to 79\% compared to the baseline FEEL algorithm while achieving similar accuracy. To the best of our knowledge, this work is the first that introduces intelligent sample exclusion during FEEL rounds. 
    
\end{itemize}

The rest of this paper is structured as follows: we review recent related works in Section ~\ref{relatedwork}. Then, we introduce the system model, learning model, and computation and communication models in Section~\ref{sysmodel}. 
The problem statement is formulated in Section ~\ref{sec:Problemformulation}. 
Experimental results are presented in Section~\ref{sec:experiment}. Finally, Section ~\ref{conclusion} concludes this work and provides directions for future extensions. 

\section{Related Work}
\label{relatedwork}
Communication and energy-constrained FEEL has been investigated by several researchers in the recent literature ~\cite{sattler2019robust,lin2017deep,wang2019adaptive,zhu2019broadband,nishio2019client,yang2018federated,liu2019edge}. The authors in ~\cite{sattler2019robust} and ~\cite{lin2017deep} proposed to compress the model parameters exchanged between edge servers and devices in order to reduce the communication cost and allow more users to join the training rounds. Another method proposed by Wang \textit{et al.}~\cite{wang2019adaptive} to optimize the number of global rounds as well as the number of local updates and minimize the value of the loss function and increase the accuracy taken into consideration the communication limitation of the wireless medium. Nishio and Yonetani~\cite{nishio2019client} addressed worker selection in heterogeneous settings and proposed an approach to select the workers that provide less computation and communication time. However, they did not consider the effects of data size on the convergence rate.

Furthermore, to address the transmission latency, the Broadband Analog Aggregation (BAA) scheme has been proposed in ~\cite{zhu2019broadband}  to reduce the transmission time between edge devices and the orchestrator server, by utilizing the superposition property of wireless channels via over-the-air computation (AirComp)~\cite{yang2018federated}. 
 AirComp advocates that concurrent transmissions can diminish multiple access latency by a factor equivalent to the number of devices (e.g., 100 times for 100 devices). This is a promising solution for fast edge learning; however, the required analog modulation makes it challenging to deploy this multi-access scheme, i.e., BAA, in modern wireless systems (digital infrastructure). To tackle this challenge, Zhu \textit{et al.} ~\cite{zhu2020one} proposed a digital aggregation method over multi-access channels, coined as One-Bit Broadband Digital Aggregation (OBDA), which represents the first attempt to implement BAA using digital modulation. Specifically, OBDA integrates the digital modulation scheme (e.g., quadrature amplitude modulation or QAM) and the state-of-the-art one-bit quantizer~\cite{zhu2020one}. 
\begin{table*}[t]
\small
    \centering
    \caption{\uppercase{Relationship between our work and the recent literature}}
    \begin{tabular}{|c|p{2cm}|p{1.7cm}|p{2cm}|p{2cm}|p{2cm}|} \hline
        \textbf{Ref} & \textbf{Completion Time} & \textbf{Synchronize the updates}& \textbf{Devices Heterogeneity}  &  \textbf{Energy Budget} & \textbf{Training Algorithm}   \\ \hline
                \cite{mo2020energy}  
        & \checkmark  & \xmarklight & \checkmark & \xmarklight & \xmarklight \\ \hline
        \cite{zeng2020energy}  
        & \checkmark  & \xmarklight & \checkmark & \xmarklight & \xmarklight \\ \hline
        \cite{wang2020federated}  
        & \checkmark  & \xmarklight & \checkmark & \xmarklight & \xmarklight \\ \hline
              Our work 
        & \checkmark  & \checkmark  & \checkmark & \checkmark & \checkmark  \\ \hline
      
    \end{tabular}
    \label{tab:related}
\end{table*}
Focusing on energy-constraints, Wang \textit{et al.} ~\cite{wang2020federated} considered the problem of energy-efficient communication and local computation resource allocation over wireless channels. They formulated an optimization problem to minimize the completion time, energy consumed during the local computation, and transmission
energy. Furthermore, the authors in ~\cite{mo2020energy} proposed an approach to minimize the total energy consumption across all workers during predefined training time. They also formulated and solved the resulting optimization problem using numerical methods.

Despite such research efforts, there is a lack in designing an energy-efficient FEEL system, as illustrated in Table \ref{tab:related}, considering optimizing the learning algorithm and the available communication computation edge network resources. This gap motivates us to propose a novel energy-efficient FEEL algorithm.
\section{System Model}
\label{sysmodel}
\begin{figure}[t]
\centering
  \includegraphics[width=0.8\linewidth]{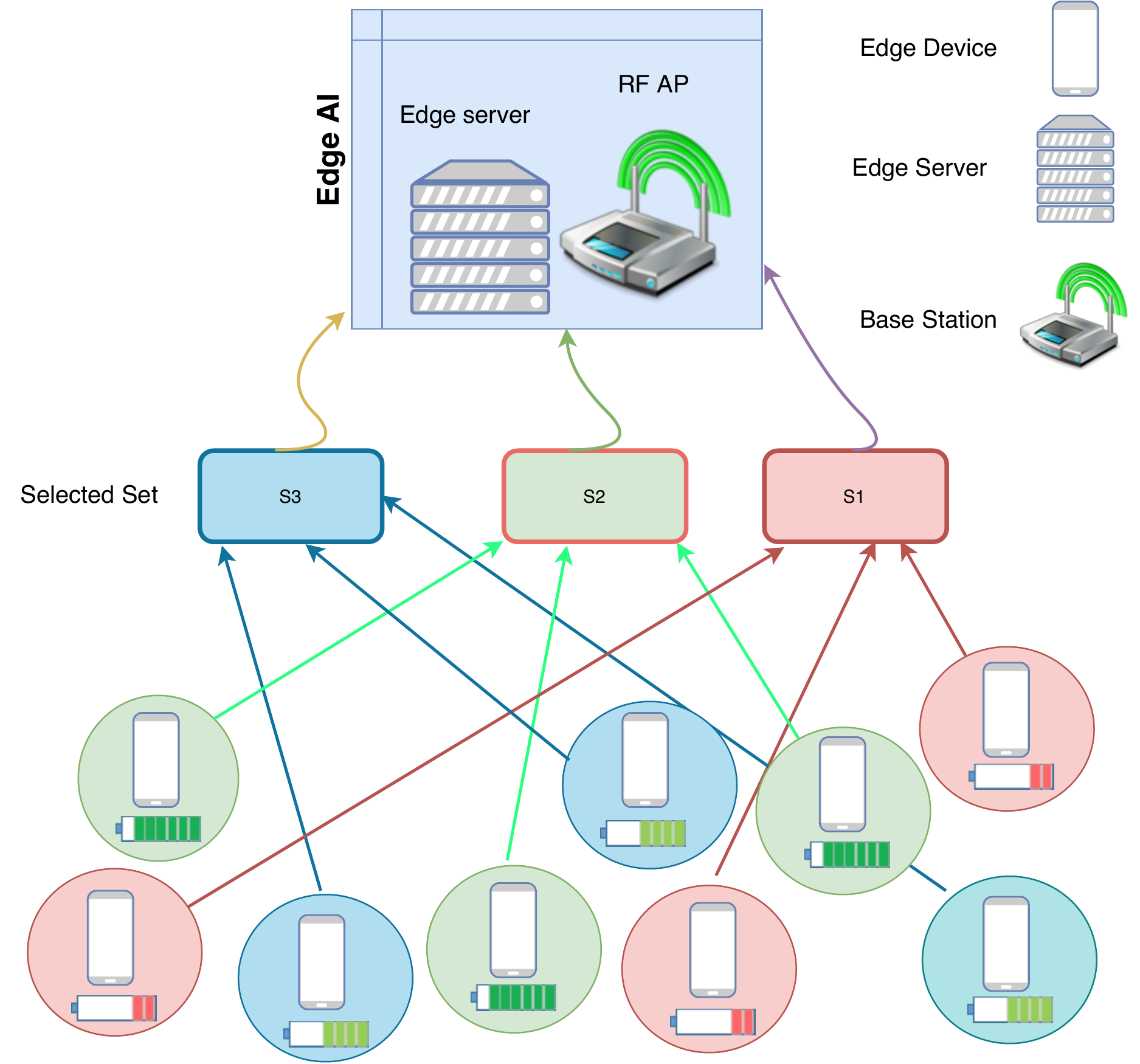}
\caption{The system model when a set of edge devices are connected to RF BS.}
\label{fig:sysmodel}
\end{figure}
As depicted in Fig.~\ref{fig:sysmodel}, the system model used in this work consists of a set of edge devices $\mathcal{K}$ connected to a BS with M-antennas that coordinates the workers to train a general global model to be used for future inference. Each worker $k \in \mathcal{K}$ has its own data $\mathcal{D}_k$ that is used to train its local model and send the update $\mathcal{\theta}_k$ back to the server, where $\mathcal{D}_k=\{{x}_{k,d}\in \mathbb{R}^d, y_{k,d} \in\mathbb{R}\}$, and $|\mathcal{D}_k|$ is the number of local data samples. 
${x}_{k,d}$ is the $d$-dimensional input data vector at  the $k$-\textit{th} worker, and $y_{k,d}$ is the corresponding label associated with ${x}_{k,d}$.  In return, the server collects all the workers' updates and averages them to form a new global model. In the beginning, the server selects a subset of workers $S$ and then sends random initialization parameters $\mathcal{\theta}_0$ so that the chosen workers can work accordingly. The selected workers use the received global model as reference when training local models to control the global and local models' divergence. After that, the server sends the updated model parameters $\mathcal{\theta}_r$, which results from averaging the whole updates at the $r$-\textit{th} round. To involve a specific worker in the training rounds, energy budget, computation, and communication capabilities are considered to establish a robust FEEL system that avoids losing selected worker updates due to insufficient energy or waiting a long time for stragglers. 

\subsection{FEEL model}
\label{sec:loss}
The local loss function captures the error of the model on the local dataset $\{{x}_{k,d},{y}_{k,d}\}$ for the $k$-\textit{th} worker at the $r$-\textit{th} round, and it is defined as follows:
\begin{equation}
F_k^{r}(\mathbf{\theta}) \triangleq \frac{1}{\left|\mathcal{D}_{k}\right|}\sum_{s\in\mathcal{D}_{k}}f_{s}(\mathbf{\theta}).
\label{eq:localLossFuncAllSamples}
\end{equation}
where $f_{s}$ captures the error of each local data sample. The total data across the edge network can be defined as follows:
$D\triangleq\sum_{k=1}^{K}|\mathcal{D}_{k}|,$
and the weight of the local data samples at the $k$-\textit{th} worker $\delta_k$ can be defined as follows: $\delta_k = \frac{|\mathcal{D}_{k}|}{D}.$
 
To train its local model, the $k$-\textit{th} worker runs its local solver, such as stochastic gradient descent (SGD), locally to minimize the loss function defined in Eq. \eqref{eq:localLossFuncAllSamples} for several local epochs denoted by $\varepsilon$. Specifically, the local model parameters  $\theta_k$ are updated as follows:
    \begin{equation}
    \label{eq:local}
    \theta_k(n)= \theta_k(n-1)-\eta \nabla F_k^{r}(\theta_k(n))
\end{equation}
where $n = 1,2,\dots, \mathbf{\varepsilon}$ is the number of local updates performed by the $k$-\textit{th} worker and $\eta$ is the step size (i.e., learning rate) at each round, $\theta_k(0)$ denotes the initial global parameters received from the server and $\theta_k(\mathbf{\varepsilon})$ denotes the last local parameters update the $k$-\textit{th} worker sends back to the server after $\mathbf{\varepsilon}$ rounds. For global loss function, after uploading all local model updates computed using \eqref{eq:localLossFuncAllSamples} and \eqref{eq:local}, the weighted global loss function across workers at the $r$-\textit{th} round is computed as:
\begin{equation}
F_r(\mathbf{\theta}) \triangleq {\sum_{k=1}^{K}\delta_k F_k^{r}(\mathbf{\theta})}.
\label{eq:globalLossFuncAllSamples}
\end{equation}
Accordingly, the global model parameters are computed as follow:
\begin{equation}
\mathbf{\theta_r}={\sum_{k=1}^{K}\delta_k \mathbf{\theta}_{k}}\label{eq:globalAverage}.
\end{equation}
$F_r(\mathbf{\theta})$ and $\mathbf{\theta_r}$ are sent to all selected workers to be used as a reference in the ($r+1$)-\textit{th} round when updating the model parameters. Thus, the aim is to find $\mathbf{\theta}^{*}$ so as  to minimize $F(\mathbf{\theta})$
\begin{equation}
\mathbf{\theta}^{*} \triangleq \arg\min F(\mathbf{\theta}).
\label{eq:learningProblem}
\end{equation}

\subsection{Local Computation model}
As stated before, the $k$-{th} worker holds $|D_k|$ data samples. To train its local model, the $k$-{th} worker splits the data $D_k$ into batches of size $b$ and trains its model for a number of epochs $\varepsilon$. Thus, the local computation time $T^{cmp}_k$ can be defined as:
\begin{align}
\label{eq:localtime2}
T^{cmp}_k =\varepsilon \frac{|D_k| \Phi}{f^\mathrm{cmp}_k}
\end{align}
where $f^\mathrm{cmp}_k$ denotes the local CPU frequency, and $\Phi$ denotes the number of cycles required  to process one sample. Namely, the server sets a deadline $\textbf T$ to synchronize the updates and avoid long waiting times especially for stragglers (i.e., the devices with low battery, low CPU speed, and bad channels). Thus, the $k$-{th} worker has to accomplish its computation and communication phases within $\textbf T$ to ensure the update synchronization. Accordingly, the local computation time should satisfy this condition:
\begin{equation}
\label{eq:localtimelocal1}
 T^\mathrm{ cmp}_k = \textbf T-  T^\mathrm{ up}_k
\end{equation}
where $T^\mathrm{ up}_k$  denotes the uploading time of the update to the server.
\subsection{Local Energy consumption model:}
From~\eqref{eq:localtime2} the corresponding local energy consumption for every $k$-\textit{th} worker due to local model training is defined as:
\begin{equation}
    \label{eq:LocalEnergy}
   E_k^{cmp} = \frac{\alpha_k}{2} (f^\mathrm{cmp}_k)^3 T^{cmp}_k
\end{equation}
where $\frac{\alpha_k}{2}$ is the energy capacitance coefficient of  $k$-\textit{th} device. 
Substituting \eqref{eq:localtime2} into the right hand-side of \eqref{eq:LocalEnergy} yields: 
\begin{align}
    \label{eq:LocalEnergy2}
   E_k^{cmp} = \frac{\alpha_k}{2} (\varepsilon (f^\mathrm{cmp}_k)^2  {|D_k| \Phi})  
\end{align}
\subsection{Radio Frequency Uploading model}
We consider Time Division Multiple Access (\emph{TDMA}) for uploading local models. We denote the uplink channel gain between the $k$-{th} worker and the $M$-antenna BS by ${\mathbf h}_k\in \mathbb C ^{M}$.
Accordingly, for a given upload interval $T^\mathrm{up}_k$, the uplink data rate achieved by the $k$-{th} worker can be defined as:
\begin{footnotesize}
\begin{align}\label{eq:achievedrate}
{R_{k}^{up} = T^\mathrm{ up}_k B~\text {log}_2\left(1 + \frac{ \left| {  {\mathbf h}_k^H {\mathbf w}_k} \right|^2 P^{ up}_k}{ {\mathbf w}_k^H \left( \sum\limits_{k' \ne k } {\mathbf h}_{k'} {\mathbf h}_{k'}^H +  \sigma^2_{0}  {\mathbf I} \right) {\mathbf w}_k }\right)},
\end{align}
\end{footnotesize} where $B$ is the bandwidth, ${\mathbf w}_k \in \mathbb C ^{M}$ denotes the received beamforming vectors from $M$-antenna BS, $P^{ up}_k$ is the $k$-\textit{th} worker transmit power, (.)$^H$ stands for the Hermitian operation, $\sigma^2_{0}$ is the spectral density power of the additive white Gaussian noise (AWGN), and ${\mathbf I}$ is the identity matrix. 
By letting $\Gamma_k = B ~\text {log}_2\left(1 + \frac{ \left| {  {\mathbf h}_k^H {\mathbf w}_k} \right|^2 P^{ up}_k}{ {\mathbf w}_k^H \left( \sum\limits_{k' \ne k } {\mathbf h}_{k'} {\mathbf h}_{k'}^H +  \sigma^2_{0}  {\mathbf I} \right) {\mathbf w}_k }\right)$, the upload latency is defined as:
\begin{align} \label{eq:uploadlatency}
 T^\mathrm{ up}_k = \frac{\xi}{\Gamma_k}
\end{align}
where $\xi$ denotes the model size. Furthermore, the transmission energy consumption of  the $k$-{th} worker is defined as: 
\begin{align}\label{eq:uploading}
E^\mathrm{ up}_{k} = T^\mathrm{up}_k P^{ up}_k.
\end{align}
\section{Problem Formulation}
\label{sec:Problemformulation}
The aim of this work is to minimize the total energy consumed during FEEL rounds, subject to constraints on the energy consumed for computation and communication and model update and upload latencies. Particularly, to keep the model updates consistent,  the selected workers should meet all required constraints. Consequently, we can formulate the optimization problem as follows:

\begin{subequations}\label{eq:OptmizedProblem1}
\begin{align}
	\textbf{P$_1$:} \quad   \underset{P^{ up}_k, T^\mathrm{ up}_k, T^\mathrm{ cmp}_k,  \atop f^\mathrm{cmp}_k, {\mathbf w}_k} \min \quad & 
\SumNoLim{r=1}{R} \SumNoLim{k}{K} \mathbf{I(k)}(E_k^{cmp} + E^\mathrm{ up}_{k} )   \label{eq:OptmizedProblema} 
\\
\text{s.t.:}  			
			\quad & E^\mathrm{cmp}_k + E^\mathrm{ up}_k \le   {E}_{k}, \quad (\forall k) \label{eq:OptmizedProblemEnergy}
\\
			\quad & T^\mathrm{ cmp}_k + T^\mathrm{ up}_k = \mathbf T, \quad (\forall k) \label{eq:OptmizedProblemDeadline}
\\
			\quad & P^\mathrm{min}_k \le P^{ up}_k \le P^\mathrm{max}_k, \quad (\forall k)	\label{eq:OptmizedProblem_pwr_transmit}
\\	
			\quad & f^\mathrm{ min}_k \le f^\mathrm{cmp}_k \le f^\mathrm{ max}_k, \quad (\forall k) \label{eq:OptmizedProblemEnergyPU}
\\
	\quad & R_{k}^{up} \ge \xi, \quad (\forall k) \label{eq:OptmizedProblem_up_model}
\\
			\quad & \left| {\mathbf w}_k \right|^2 = 1, \quad (\forall k) \label{eq:OptmizedProblemBeamForming}
\end{align}
\end{subequations}
where $\mathbf{I(k)}$ is an indicator function that specifies whether the k-\textit{th} worker is involved in the $r$-\textit{th} round or not. Constraint \eqref{eq:OptmizedProblemEnergy} ensures that the energy consumed for computation and communication does not exceed the energy budget of the $k$-\textit{th} worker. 
The constraint  \eqref{eq:OptmizedProblemDeadline} is set to guarantee that the total computation and upload time is restricted to round deadline ${\mathbf T}$ to synchronize the updates.
The transmit power of every selected worker is restricted in \eqref{eq:OptmizedProblem_pwr_transmit} to be between the minimum transmit power $ P^\mathrm{min}_k$ and the maximum transmit power $ P^\mathrm{max}_k$. Constraint ~\eqref{eq:OptmizedProblemEnergyPU} ensures that the CPU-frequency of the $k$-\textit{th} worker ranges between the minimum $f^\mathrm{ min}_k$ and maximum $f^\mathrm{ max}_k$ CPU frequencies.
Constraint \eqref{eq:OptmizedProblem_up_model} ensures that the upload time of the $k$-\textit{th} worker is sufficient to send the model (i.e., the updated parameters) $\xi$ to the server.
Last, constraint \eqref{eq:OptmizedProblemBeamForming} ensures that the received beamforming vectors have direct direction to the M-antenna BS. 

Evidently, \textbf{P$_1$} is intractable as it requires the availability of future information about the participating workers and their channels and mobility. Also, variables $P^{ up}_k$, $\mathrm{ T^\mathrm{ up}_k}$, and $T^\mathrm{ cmp}_k $ are all coupled in constraints \eqref{eq:OptmizedProblem_up_model}, \eqref{eq:OptmizedProblemEnergy}, and \eqref{eq:OptmizedProblem_up_model}. 

It is worth noting that the optimal beam vector ${\mathbf w}_k$ can be defined as~\cite{tran2020lightwave}:
\begin{align}
{\mathbf w}^{\star}_j =  \arg \underset{\left| {\mathbf w}_k \right|^2 = 1} \max R_{k}^{up} \quad (\forall k).
\end{align}
According to Rayleight-Ritz quotient \cite{Parlett,tran2020lightwave}, ${\mathbf w}^{\star}_k$ can be obtained by finding the eigenvector corresponding to the largest eigenvalue of the matrix $ {\mathbf h}_k{\mathbf h}_k^H \left( \sum_{k' \ne k } {\mathbf h}_{k'} {\mathbf h}_{k'}^H +  \sigma^2_{0}  {\mathbf I} \right)^{-1}$.  
\section{Proposed Approach}
\label{sec:proposedsol}
To solve \textbf{P$_1$}, we propose a novel three-phase heuristic. In the first phase of the heuristic, a novel local training algorithm is introduced to allow workers only to include the samples that provide significant enhancement for the global model and, as a consequence, reduce the local computation energy. The details of the proposed algorithm are presented in Section \ref{sec:proposedsol}. In the second phase, we tune the transmit time, transmit power, and local CPU speed based on the "reduced" local samples. Finally, in the third phase, we utilize the approach presented in ~\cite{tran2020lightwave} to find the optimal value of ${\mathbf w}_k$ which in return maximizes $R_{k}^{up}$ to reduce the upload time and the corresponding transmit power. In our approach, all selected workers receive the global parameters from the edge server and use their entire local samples to update the received model parameters only once (i.e., initialization epoch $\varepsilon$) in order to specialize the global parameters and reduce the divergence between the global and local models. Then, workers use the updated model to predict the local samples and exclude the samples classified with a probability higher than a preset threshold $\vartheta$. The threshold specifies below which samples are excluded in future epochs. Given the number of local samples $|D_k|$, the number of excluded samples is denoted as $\iota$. Therefore, the number of samples included in future epochs is $|D_k|- \iota$. Accordingly, the computation time needed to complete the update task can be rewritten as:
\begin{align}
\label{eq:localtimeproposed}
T^{cmp}_k= (\varepsilon-1)  \frac{(|D_k|-\iota) \Phi}{f^\mathrm{cmp}_k} + \frac{|D_k| \Phi}{f^\mathrm{cmp}_k}  \nonumber \\
= \frac{(\varepsilon \phi |D_k|) - \iota (\varepsilon-1)}{f^\mathrm{cmp}_k}
\end{align}
Consequently, the corresponding total energy consumed for local computation can be defined as:
\begin{align}
    \label{eq:LocalEnergyProposed}
   E_k^{cmp} = \frac{\alpha_k}{2} (\varepsilon-1) (f^\mathrm{cmp}_k)^2  {(|D_k|-\iota) \Phi}) +   \frac{\alpha_k}{2} (f^\mathrm{cmp}_k)^2  {|D_k| \Phi}) 
\end{align}
In Algorithm ~\ref{alg:Proposedalg1}, the server initiates the global model parameters and determines the threshold probability that is used to select the samples to be included in the local training.  The selected samples with $P \leq \vartheta$ are used to train the local model for the rest of the epochs. The steps of these algorithms are summarized in  Algorithms ~\ref{alg:Proposedalg1} and ~\ref{alg:Proposedalg2}.
\begin{algorithm} []
    \caption{FEEL Algorithm} \label{alg:Proposedalg1}
       \textbf{Initialization:}$\theta_{0}$, $e>0$, $b$ \;
        \For{$r=1$ to $R$}
        {
        \textbf{worker Selection:} Server selects a subset of available devices based on the existing resource budget to train the model.\;
      \textbf{Local training:} Each worker $k$ receives $\theta_{r-1}$ and $\nabla F_{r-1}(\theta)$  from the server, and then uses Algorithm ~\ref{alg:Proposedalg2} to train its local model\;
   {\textbf{Communication Phase:} Each worker $k$ sends back $\theta_k$ and $\nabla F_k(\theta)$, $\forall k$, to the edge server}\;  \label{commround}
\textbf{Aggregation and Returns:} The edge server aggregates all updates and then modify the global model parameters $\mathbf{\theta_r}$ and $\nabla F_r(\mathbf{\theta})$ and then returns-back the new parameters to all workers\; \label{line:bs}
       }
\end{algorithm}
\setlength{\textfloatsep}{0pt}
\begin{algorithm} []
    \caption{Local Energy-Efficient Algorithm} \label{alg:Proposedalg2}
    \textbf{Local Predicting:} Each worker $k$ uses the updates model in the first epoch and predict all local samples;
     \textbf{Set $\mathcal{D}_k^{r} = \{ \}$}\;
     \For{$d = 1$ to $|D_k|$}{
     \If{$P({x}_{d}) \leq \vartheta$}  
     { $\mathcal{D}_k^{r} = \mathcal{D}_k^{r} \cup\{{x}_{d} , {y}_{d}$\}
     }
     }
      \For{$epoch=2$ to $\varepsilon$}{
      Each worker $k$ trains its local model using only $\mathcal{D}_k^{r}$
      }
\end{algorithm}
\setlength{\textfloatsep}{0pt}
Next let $\rho = (\varepsilon \phi |D_k|) - \iota (\varepsilon-1) $, then considering \eqref{eq:LocalEnergy} and \eqref{eq:OptmizedProblem_pwr_transmit}, constraint \eqref{eq:OptmizedProblemDeadline} can be rewritten as
\begin{align}\label{eq:OptmizedProblem_pwr_transmit1}
0 <  \mathbf T - \frac{\rho}{f^\mathrm{min}_k} \leq  T^\mathrm{ up}_k  \leq  \mathbf T - \frac{\rho }{f^\mathrm{max}_k} \leq \mathbf T. \quad (\forall k)
\end{align}
Further, let $\beta_k = \frac{ \left| {  {\mathbf h}_k^H {\mathbf w}_k^{\star}} \right|^2 }{ {\mathbf w}_k^{\star H} \left( \sum\limits_{k' \ne k } {\mathbf h}_{k'} {\mathbf h}_{k'}^H +  \sigma^2_{0}  {\mathbf I} \right) {\mathbf w}_k^{\star} }$. Hence from \eqref{eq:achievedrate} and by using exponent of log rule, \eqref{eq:OptmizedProblem_up_model} can be derived as:
\begin{align}\label{eq:local_trans}
P^{ up}_k  = \frac{2^{\frac{\xi}{T^\mathrm{ up}_k B}}-1}{\beta_k}.
\end{align}
By substituting \eqref{eq:local_trans} into the right hand side of \eqref{eq:uploading}, we have:
\begin{align}
\label{eq:energyUploading2}
E^\mathrm{up}_k= T^\mathrm{ up}_k \frac{2^{\frac{\xi}{T^\mathrm{ up}_k B}}-1}{\beta_k}.
\end{align}
$T^\mathrm{ up}_k$, $T^\mathrm{cmp}_k$ and $f^\mathrm{cmp}_k$ should be tuned to synchronize the workers and minimize the total energy consumption of the FEEL system. From~\eqref{eq:OptmizedProblem_pwr_transmit}, we can infer that $T^\mathrm{ up}_k$ is bounded and the Golden-section search method can be used to find its optimal value~\cite{William2007, tran2020lightwave} as follows: 
\begin{center}
\begin{tabular}{|l|}
\hline
If $E(a_{i+1}) \le  E(b_{i+1})$ \tabularnewline
\quad   $T^\mathrm{ up}_k \in [ a_i, b_{i+1}]$ \tabularnewline
Else\tabularnewline
 \quad  $T^\mathrm{ up}_k \in [ a_{i+1}, b_{i}]$.\tabularnewline
\hline
\end{tabular}
\end{center}
In view of this, $a_{i+1} = a_i + \varphi (b_i - a_i)$, $b_{i+1} = a_i + (1-\varphi) (b_i - a_i)$, $\varphi = \frac{3-\sqrt{5}}{2}$, $a_0 = \mathbf T - \frac{\rho}{f^\mathrm{min}_k} $, and $b_0 =  \mathbf T - \frac{\rho}{f^\mathrm{max}_k}$ ~\cite{William2007}. The optimal solution can be found by iterativly shortening the interval between the upper-bound and lower-bound  using the golden ratio $\varphi$. The solution can be easily attained and the used method ensures convergence. Then, $T^\mathrm{cmp}_k$ can be calculated using \eqref{eq:localtimelocal1}. Finally, $f^\mathrm{cmp}_k$ is computed \eqref{eq:OptmizedProblemEnergyPU}:
\begin{align}\label{eq:local_time_updated}
f^\mathrm{cmp}_k = \frac{(\varepsilon \phi |D_k|) - \iota (\varepsilon-1)}{T^\mathrm{cmp}_k}
\end{align}
\section{Simulation and Numerical Results}
\label{sec:experiment}
In our experiments, we consider a FEEL environment as in ~\ref{fig:sysmodel} with a total bandwidth of $B=1 {MHz}$, and noise power $\sigma^2 = 10^{-6}$. The distance between the edge workers and the BS is uniformly distributed between $5 m$ and $20 m$. For the wireless channel model, we use Rician distribution with a Rician factor of 8 dB and a path loss exponent factor of $3.2$. The number of antennas is $m = 8$ for the BS and $m = 1$ for each of the $k$ workers. The maximum and minimum transmit powers are set to $P_{max} = 20$ dBm and $P_{min} = −10$ dBm, respectively. We use the MNIST datasets under a realistic federated setting, imbalanced and non-i.i.d data distribution with $1000$ users, and different convolutional neural networks (CNN) models. We utilize the mini-batch SGD as a local solver and evaluate the global model every round. The data is split into $80\%$ for training and $20\%$ for testing.   

\begin{figure}[t]
\centering
		\includegraphics[width=0.8\linewidth]{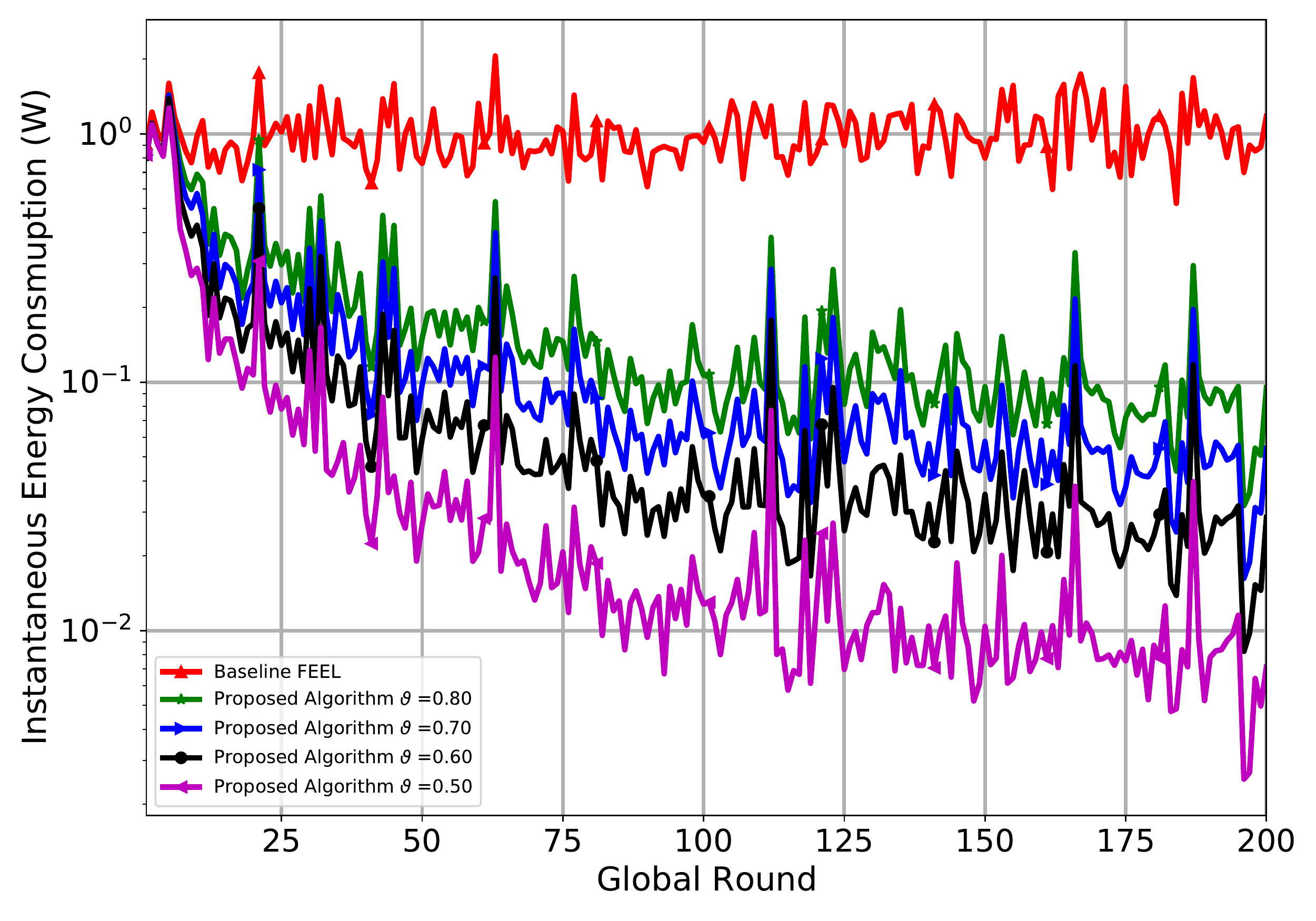}
	\caption{Instantaneous Energy Consumption when the number of global rounds is 200.}
		\label{F:EnergyConsumption200}
\end{figure}

\begin{figure}[t]
\centering
		\includegraphics[width=0.8\linewidth]{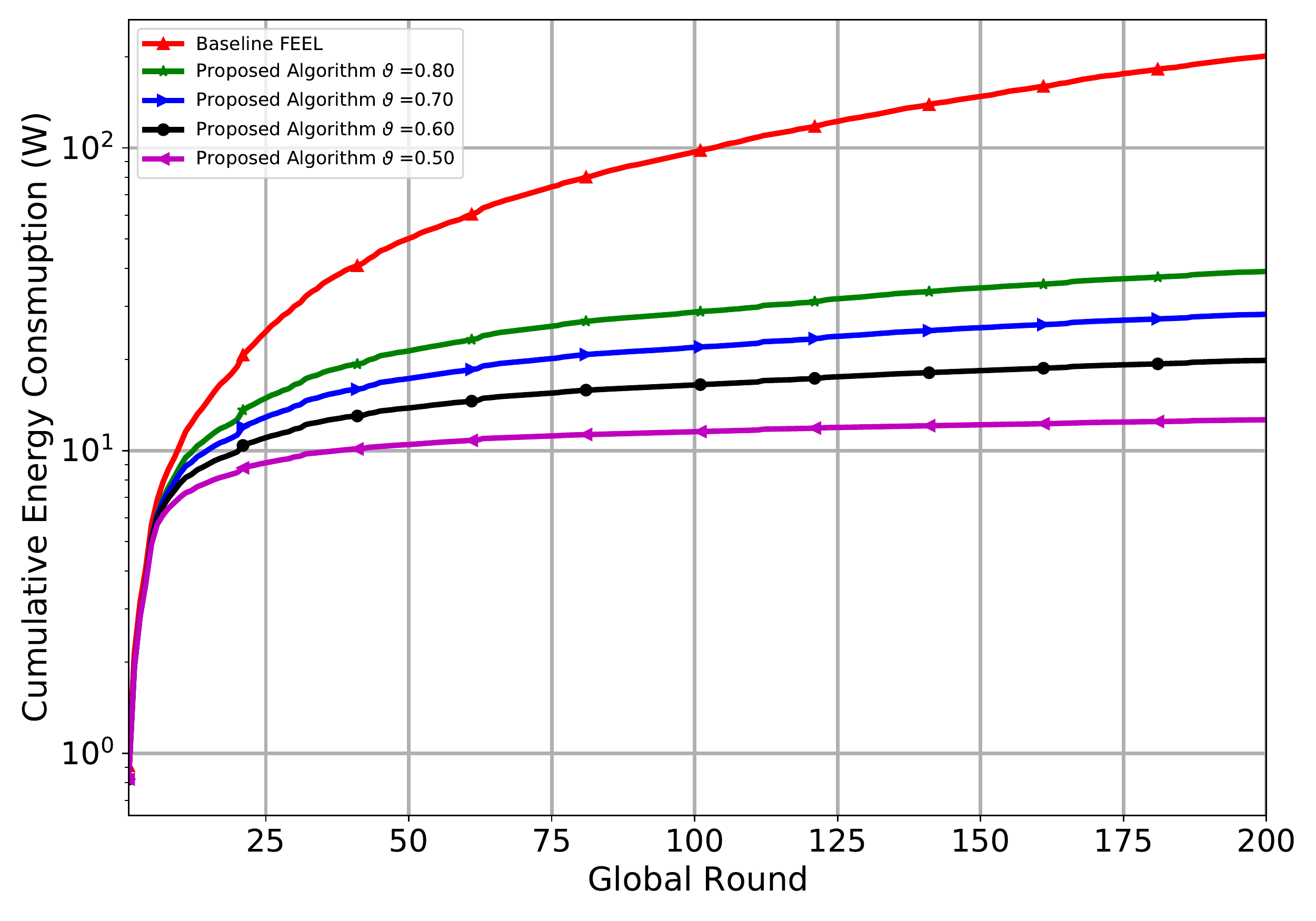}
	\caption{Cumulative Energy Consumption when the number of global rounds is 200.}
		\label{F:EnergyConsumption200_cum}
\end{figure}

\Cref{F:EnergyConsumption200,F:EnergyConsumption200_cum} show the energy consumption during FEEL rounds when the number of global rounds is $200$. It is evident that the proposed algorithm substantially reduces the total consumed energy. This stems from excluding the samples having less impact on the model training, affecting computation and communication energy consumption. For computation, only a subset of the workers' samples is injected into training for $(\varepsilon-1)$ epochs. 
In contrast, the baseline FEEL algorithm consumes more energy as all local samples are included in $\varepsilon$ local iterations, which affects the transmission power.   

\begin{figure}[!t]
	\centering
		\includegraphics[width=0.8\linewidth]{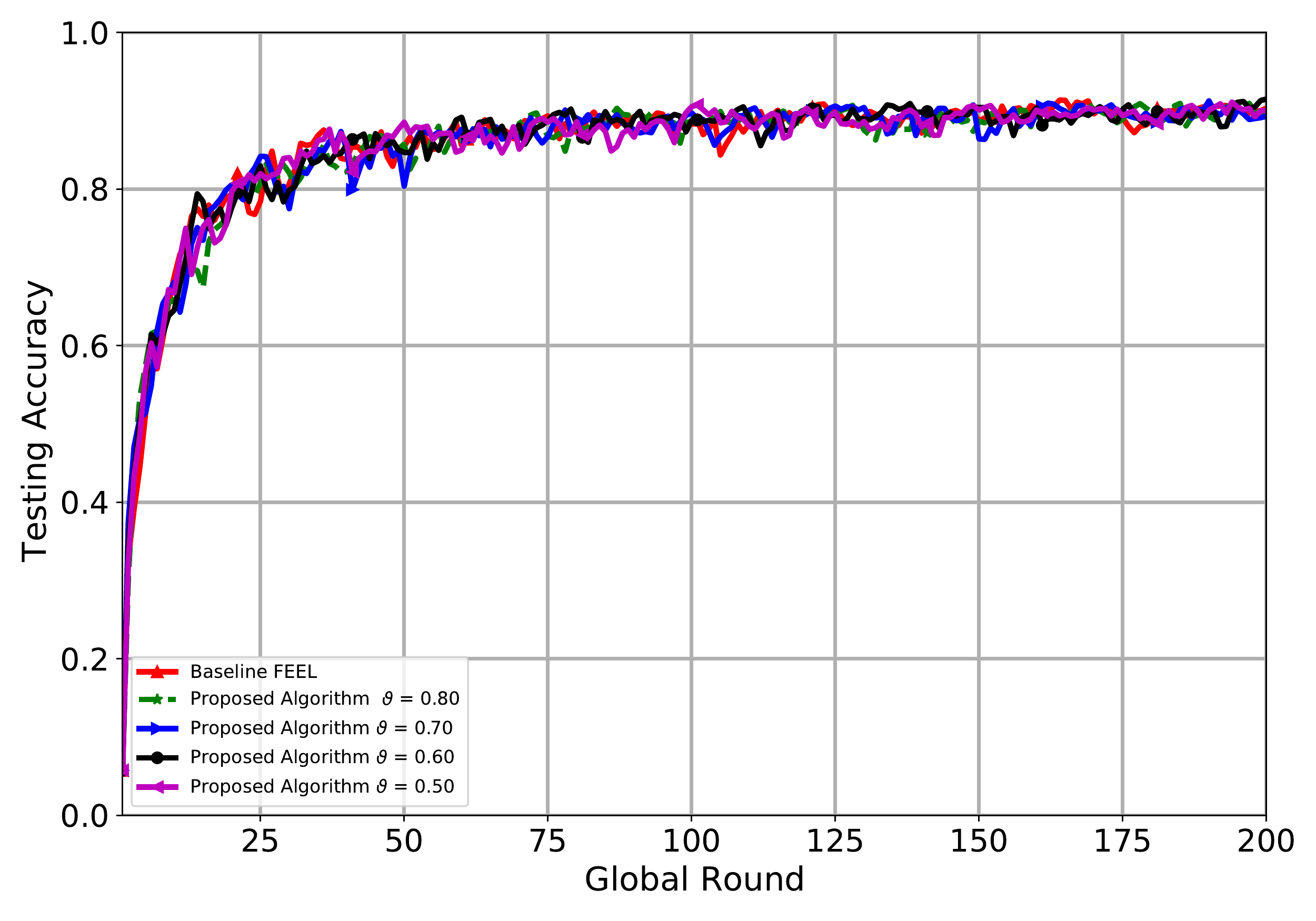}
	\caption{Testing Accuracy (MNIST) when the number of global rounds is 200.}	\label{F:mnist_accu200}
\end{figure}

\begin{figure}[!t]
	\centering
		\includegraphics[width=0.8\linewidth]{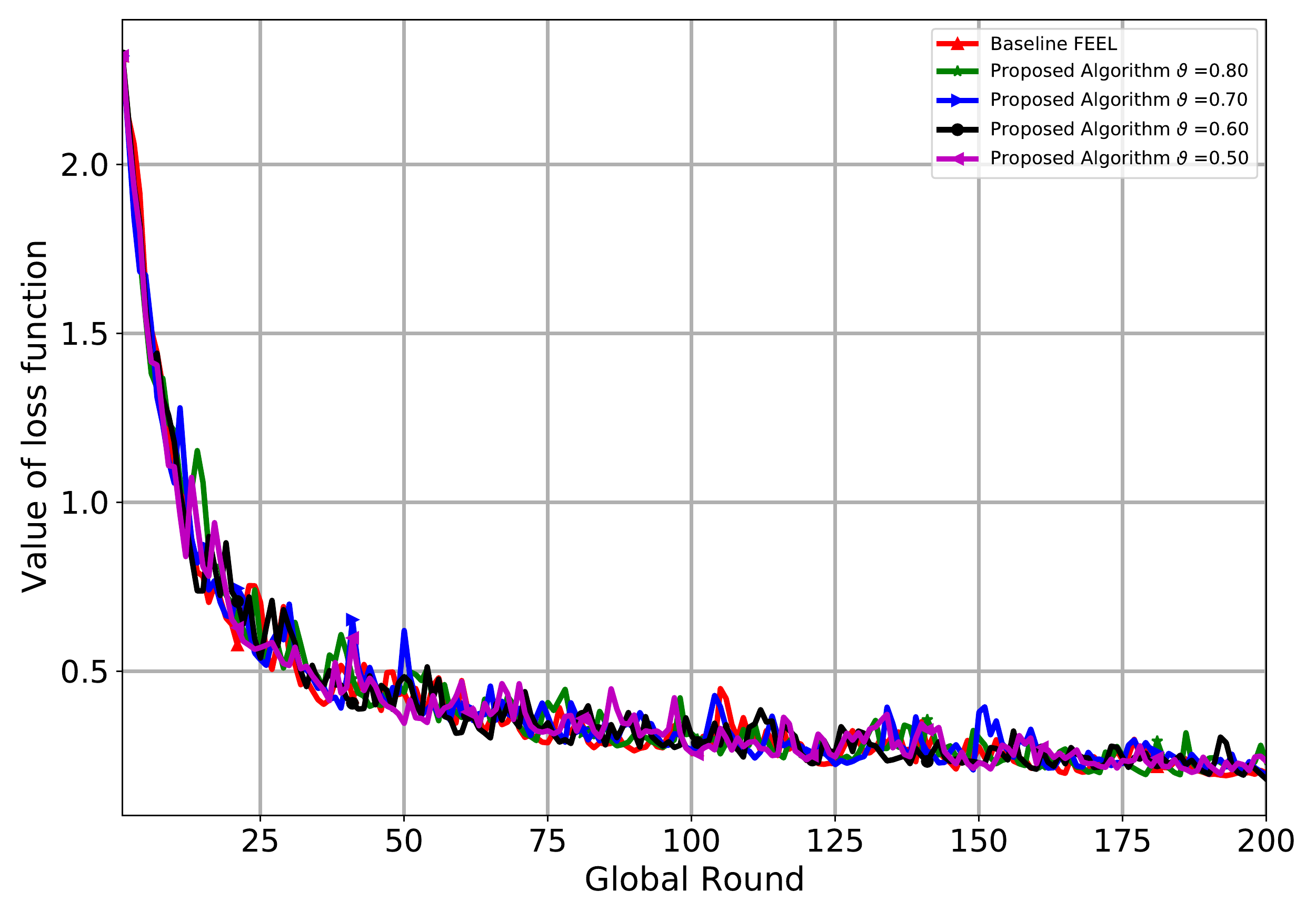}
	\caption{Value of the loss function (MNIST) when the number of global rounds is 200.}
		\label{F:mnist_loss200}
\end{figure}

\Cref{F:mnist_accu200,F:mnist_loss200} show the identification accuracy and loss of handwritten digits (MNIST) when the number of global rounds is $200$ and $\vartheta = 0.5, 0.6, 0.7,$ and $0.8$. From these figures, it is evident that the proposed algorithm provides approximately similar accuracy and loss, especially when the threshold probability is higher than $0.70$. However, both accuracy and loss worsen when the threshold probability is lower than $0.70$ as we can see when $\vartheta = 0.50$. This is because most of the excluded samples are predicted with low probability, decreasing the number of samples used to train the local models for the rest of the epochs. 
Overall, our proposed approach considers computation and communications aspects that lead to significant energy efficiency enhancements supporting synchronized FEEL systems' real-life deployments. The observed enhancements stem from our proposed methods to intelligently exclude training samples that do not significantly contribute to the global model. 

\section{Conclusion}
\label{conclusion}
In this work, a novel energy-efficient FEEL approach was proposed. Our proposed approach takes advantage of local computation and communication resources to substantially reduce the energy consumed by selected workers to train their local models.  In our proposed approach, each worker tunes the received global model parameters while intelligently excluding the predicted samples with high probability based on a predefined threshold as such samples do not significantly contribute to the learning model and can adversely impact energy consumption.  The proposed approach tunes the transmit power and local CPU speed of workers in a FEEL system to enhance energy efficiency.  Our experimental results demonstrate outstanding potential for reducing the total energy consumption of FEEL systems. Eventually, we show that energy consumption can be reduced by optimizing the available computation and communication resources and designing efficient local training algorithms. For future work, investigating the threshold's optimal value should be considered for a more efficient local training algorithm.

\section*{Acknowledgement}
This publication was made possible by NPRP-Standard (NPRP-S) Thirteen (13th) Cycle grant \# NPRP13S-0201-200219 from the Qatar National Research Fund (a member of Qatar Foundation). The findings herein reflect the work, and are solely the responsibility, of the authors.
\bibliographystyle{IEEEtranTIE}
\bibliography{Ref}

\end{document}